\pdfoutput=1

\documentclass[11pt]{article}

\usepackage[final]{acl}

\usepackage{times}
\usepackage{latexsym}

\usepackage[T1]{fontenc}

\usepackage[utf8]{inputenc}

\usepackage{microtype}

\usepackage{inconsolata}

\usepackage{graphicx}

\usepackage{amsmath}
\usepackage{url}
\usepackage{fixltx2e}
\usepackage{multirow}
\usepackage{booktabs}

\title{Generate, Discriminate, Evolve: Enhancing Context Faithfulness via Fine-Grained Sentence-Level Self-Evolution}

\author{First Author \\
  Affiliation / Address line 1 \\
  Affiliation / Address line 2 \\
  Affiliation / Address line 3 \\
  \texttt{email@domain} \\\And
  Second Author \\
  Affiliation / Address line 1 \\
  Affiliation / Address line 2 \\
  Affiliation / Address line 3 \\
  \texttt{email@domain} \\}

\author{Kun Li$^{\heartsuit}$\thanks{$\;\;$Equal contribution.}$\,$, Tianhua Zhang$^{\heartsuit*}$, Yunxiang Li$^{\heartsuit}$, Hongyin Luo$^{\diamondsuit}$, \\ \bf
Abdalla Moustafa$^{\heartsuit}$, Xixin Wu$^{\heartsuit}$, James Glass$^{\diamondsuit}$, Helen Meng$^{\heartsuit}$ \\
$^\heartsuit$The Chinese University of Hong Kong, Hong Kong SAR, China \\
$^\diamondsuit$Massachusetts Institute of Technology, Cambridge MA, USA \\
\texttt{kunli@se.cuhk.edu.hk, thzhang@link.cuhk.edu.hk}
}

\begin{document}
\maketitle
\begin{abstract}
Improving context faithfulness in large language models is essential for developing trustworthy retrieval augmented generation systems and mitigating hallucinations, especially in long-form question answering (LFQA) tasks or scenarios involving knowledge conflicts. Existing methods either intervene LLMs only at inference without addressing their inherent limitations or overlook the potential for self-improvement. In this paper, we introduce \text{GenDiE} (\textbf{Gen}erate, \textbf{Di}scriminate, \textbf{E}volve), a novel self-evolving framework that enhances context faithfulness through fine-grained sentence-level optimization. \text{GenDiE} combines both generative and discriminative training, equipping LLMs with self-generation and self-scoring capabilities to facilitate iterative self-evolution. This supports both data construction for model alignment and score-guided search during inference. Furthermore, by treating each sentence in a response as an independent optimization unit, \text{GenDiE} effectively addresses the limitations of previous approaches that optimize at the holistic answer level, which may miss unfaithful details. Experiments on ASQA (in-domain LFQA) and ConFiQA (out-of-domain counterfactual QA) datasets demonstrate that \text{GenDiE} surpasses various baselines in both faithfulness and correctness, and exhibits robust performance for domain adaptation.
\end{abstract}

\section{Introduction}
Large language models have achieved remarkable success across various natural language processing tasks \citep{openai_blog_gpt4o_2024, anthropic_blog_clause_3.5_sonnet_2024, deepseekai2025deepseekr1incentivizingreasoningcapability}. Despite their impressive performance, LLMs are prone to \textit{hallucinations}---generating plausible yet nonfactual information \citep{zhang2023sirenssongaiocean, li2024decodinggraphsfaithfulsound}. This limitation poses critical risks in domains where accuracy and reliability are paramount. Retrieval-Augmented Generation (RAG) has emerged as a promising framework to mitigate hallucinations and enhance context-faithfulness \citep{nguyen2024sfrragcontextuallyfaithfulllms} by grounding LLM outputs in provided knowledge \citep{gao2024retrievalaugmentedgenerationlargelanguage, 10.1145/3637528.3671470, luo2023sailsearchaugmentedinstructionlearning}. Despite its advantages, \textit{knowledge conflicts} \cite{xie2024adaptive, xu2024knowledgeconflictsllmssurvey} between LLMs' parametric memory and external context can undermine its effectiveness. LLMs may overly rely on their internal priors while disregard provided contexts, failing to meet user requirements or incorporate the latest updates \citep{jin2024tugofwarknowledgeexploringresolving, bi2024contextdpoaligninglanguagemodels}. Additionally, when tasked with \textit{long-form} generation, such as long-form question answering (LFQA) \citep{stelmakh-etal-2022-asqa, fan-etal-2019-eli5} that aims to provide in-depth and paragraph-length responses, maintaining context-faithfulness throughout the text is still challenging \citep{stolfo-2024-groundedness}. Consequently, it is crucial to develop robust mechanism to alleviate faithfulness hallucination and ensure trustworthiness.

Many recent efforts enhance context faithfulness of LLMs through inference-time interventions, such as improving prompting strategies \citep{zhou-etal-2023-context} and context-aware decoding \citep{shi-etal-2024-trusting} to increase the output probability on contextual information. While effective, these method do not fundamentally address the models' inherent limitations \citep{bi2024contextdpoaligninglanguagemodels}. Training-based approaches, including adaptive retrieval and self-critiquing generations with reflection tokens \citep{asai2024selfrag} and aligning LLMs via Direct Preference Optimization (DPO) towards faithful responses \citep{bi2024contextdpoaligninglanguagemodels}, improve faithfulness by updating model parameters. However, these approaches typically rely on one-round optimization and do not fully explore the potential of continuous refinement. Furthermore, existing methods often train models in complete answer level holistically, which can overlook unfaithful details, particularly in the cases of long-form generation \citep{lai2024stepdpostepwisepreferenceoptimization}.

To address these challenges, we introduce a novel self-evolving framework \text{GenDiE} (\textbf{Gen}erate, \textbf{Di}scriminate, \textbf{E}volve) for enhancing LLMs' context faithfulness through fine-grained, sentence-level optimization. Our framework addresses the limitations of conventional answer-level training paradigms by operating at the sentence level, treating each constituent sentence of a response as an independent optimization unit. Central to our approach is a unified training strategy that integrates generation and evaluation capabilities through self-evolution. The model learns to produce context-grounded sentences while developing discriminative capabilities to distinguish between faithful and unfaithful responses. During each training phase, \text{GenDiE}  generates candidate sentences through tree-structured sampling, self-scores their faithfulness, and constructs contrastive sentence pairs serving as training data in the next iteration. This design enables continuous model improvement through successive self-evolve cycles. For inference, we design hierarchical decoding that first generates candidate sentences through standard methods, then selects optimal outputs using the model's learned scoring capacity to fully utilize both generative and discriminative capabilities that conventional single-stage decoding neglects.

We evaluate \text{GenDiE} on two benchmarks for context-faithful generation and the results demonstrate its effectiveness. \text{GenDiE} surpasses various baselines in two dimensions, faithfulness and correctness, and exhibits robust performance even in out-of-domain settings. Remarkably, our approach enables the model to consistently improve with each successive training iteration. We further conduct comprehensive experiments to verify the effectiveness of the self-scoring function, as well as the superiority of sentence-level optimization.





In summary, we make the following contributions:
\begin{itemize}

\item We propose \text{GenDiE}, a novel self-evolving framework that addresses the critical challenge of maintaining faithfulness in LLM responses through iterative self-improvement.

\item \text{GenDiE} operates at a fine-grained sentence level, offering more precise control over faithfulness compared to previous methods that typically operate on entire response sequences. 

\item \text{GenDiE} integrates both generative and discriminative capabilities through multi-task training, enabling models to not only generate faithful responses but also effectively discriminate between faithful and unfaithful content. 

\end{itemize}

\section{Methodology}
\begin{figure*}[ht]
\includegraphics[width=1\textwidth]{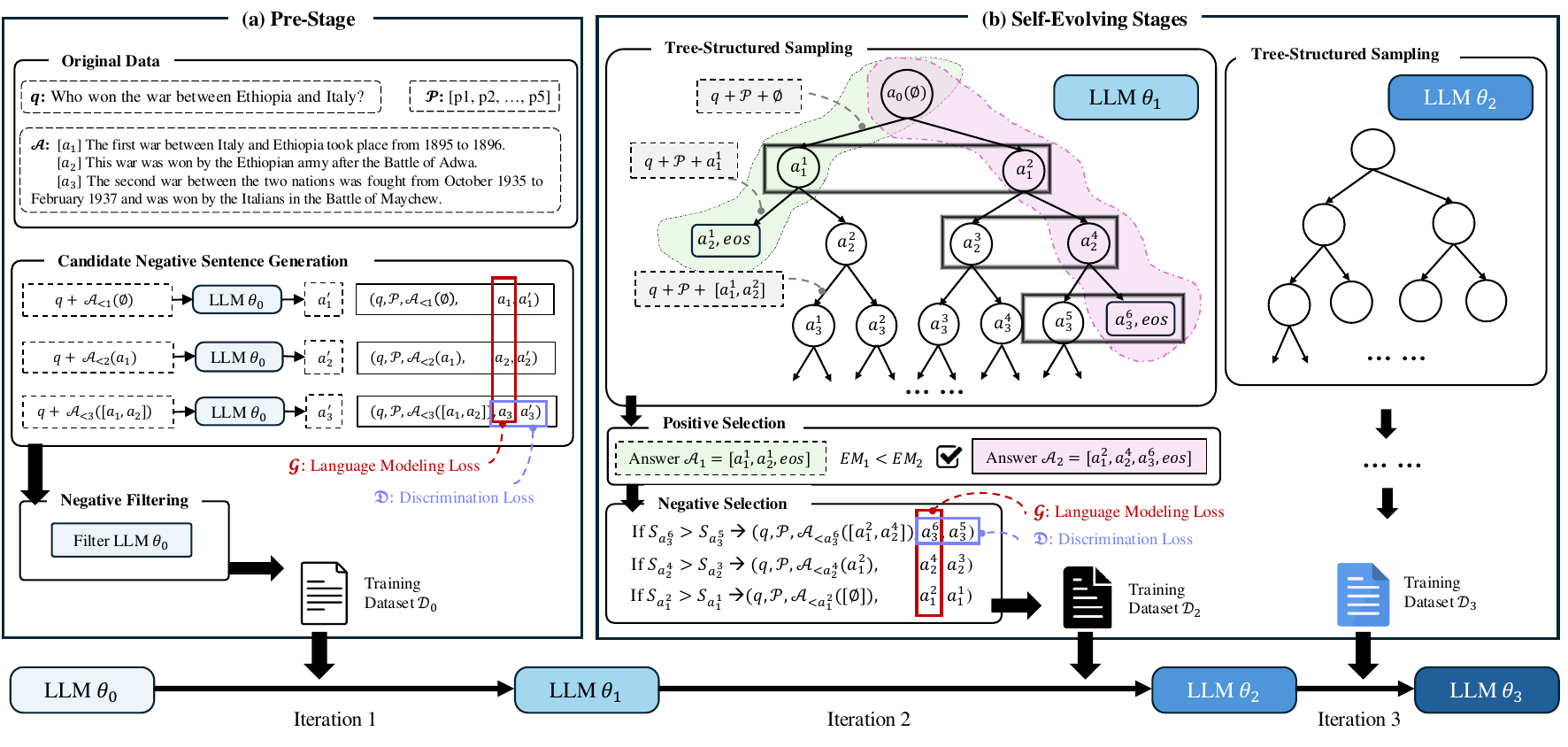}
\caption{An overview of \text{GenDiE}: (a) Pre-stage (\S\ref{sec: per-stage-data-construction}) uses gold answer sentences from a seed dataset as target faithful instances, while filtered self-generated sentences---produced without access to supporting passages---serve  as negative samples. (b) Self-evolving stages (\S\ref{sec: self-evolving}) leverage models from previous iteration for both self-generation and self-scoring, constructing training datasets via tree-structured sampling. Throughout all stages of the self-evolving framework, both \textcolor{red}{language modeling loss} (optimizing towards the target instances $a$) and \textcolor{blue}{discrimination loss} (assigning higher faithfulness scores to $a$ over $a'$) are incorporated (\S\ref{sec:training}).} 
\label{fig: overview}
\end{figure*}

\noindent\textbf{Task Formulation} 
We focus on long-form question answering (LFQA) task which requires models to generate long and detailed answers by leveraging the evidence documents provided in the input \citep{xu-etal-2023-critical}. Our goal is to train the model $M$ to generate faithful long-form answer $A=\{a_1,a_2,...a_{|A|}\}$, where $a_i$ denotes the $i$-th sentence, in response to a given input question $q$ and its corresponding evidence passages $P=\{p_j\}_{j=1}^k$. For training, we assume the availability of answer labels $a^*$ from a seed (in-domain) dataset to construct the initial training data. To comprehensively assess context faithfulness, we also employ a counterfactual dataset for out-of-domain evaluation.

\noindent\textbf{Overview} We present \text{GenDiE}, a \textit{self-evolving} framework (\S\ref{sec: self-evolving}) that enhances LLMs context-faithfulness in fine-grained sentence-level optimization (\S\ref{sec:sentence-level}), integrating both generation and discrimination capabilities. This enables LLMs to distinguish between faithful and unfaithful responses, facilitating self-scoring for both training data construction (\S\ref{sec: self-evolving stages}) and score-guided search during inference (\S\ref{sec:inference}). Our approach employs iterative self-training, where the models can self-generate, self-score, and therefore self-improve through multiple training iterations. 

\subsection{Training}
\label{sec:training}
\subsubsection{Sentence-Level Optimization}
\label{sec:sentence-level}
The majority of existing approaches train models at answer level, treating an entire answer as the training target. We believe this paradigm provides limited supervision signal for learning faithful generation, especially in LFQA task where the sentences in a lengthy answer often exhibit varying levels of faithfulness. By treating the answer as a monolithic unit, these approaches fail to capture the nuanced differences in faithfulness across individual sentences. To train a model with finer-grained supervision, we propose a sentence-level optimization. We split a target answer $A$ into a set of sentences $\{a_i\}^{|A|}_{i=1}$, and each sentence is used as a separate training instance\footnote{\noindent Answers are split based on sentence-ending punctuation marks, including periods, exclamation and question marks.}.

In addition to introducing finer-grained supervision, the sentence-level optimization can also enable sentence-level search during inference (\S\ref{sec:inference}).

\subsubsection{Self-Evolving}
\label{sec: self-evolving}
\noindent\textbf{Iterative Training for Self-evolving} We propose training the model iteratively to enable self-evolving. At $t$-iteration, the answer sentences used for training are generated and evaluated (in terms of their faithfulness) by the model $\theta_{t-1}$ obtained at $t-1$-iteration. In this way, the model gets self-evolved with progressively improved training data. 

\noindent\textbf{Training Objective} To achieve the goal of self-evolving within our single-model framework, the model should grasp the abilities of both generation and scoring in faithfulness. With this motivation, we train the model with multi-task training paradigm. Furthermore, inspired by Odds Ratio Preference Optimization \citep{hong-etal-2024-orpo}, an algorithm that optimizes language generation and preference alignment simultaneously, we train our model by maximizing 
\begin{equation}
\resizebox{1\linewidth}{!}{%
    $\begin{aligned}
    \mathcal{L}  &= \sum_{a\in A} \Biggl(\underbrace{\log\mathcal{P}_{\theta}(a|q,P,A_{\prec a})}_{\text{Language Modeling Objective}}+ \\
    &\lambda \underbrace{\log \sigma \left( \log\frac{\mathcal{P}_{\theta}(a|q,P,A_{\prec a})(1-\mathcal{P}_{\theta}(a'|q,P,A_{\prec a}))}{\mathcal{P}_{\theta}(a'|q,P,A_{\prec a})(1-\mathcal{P}_{\theta}(a|q,P,A_{\prec a}))} \right)}_{\text{Discrimination Objective}} \Biggr);
    \end{aligned}$
}
\label{eq: training_objective}
\end{equation}
where
\begin{equation}
\resizebox{1\linewidth}{!}{%
$\displaystyle
\mathcal{P}_{\theta}(a|q,P,A_{\prec a})=\prod_{t=1}^{|a|}{\mathcal{P}_{\theta}(a[t]|q,P,A_{\prec a}, a[1:t-1])}^{1/|a|}.
$%
}
\label{eq: language_loss}
\end{equation}
$a$ and $a^{'}$ in Eq.\ref{eq: training_objective} denote the target and negative sentence respectively, and they share a common prefix $A_{\prec a}$. The prefix $A_{\prec a}$ includes all the sentences in the answer $A$ preceding $a$ ($a^{'}$). $a[j]$ in Eq.\ref{eq: language_loss} denotes the $j$-th token of $a$. In Eq.\ref{eq: training_objective}, the first term in the summation is the vanilla language modeling objective, aimed to maximize the probability of generating $a$ by the model. Following \citet{zhang-etal-2024-two}, we define the faithfulness score of sentence $a$ to passages $P$ as $S_a = \text{log}\mathcal{P}_{\theta}(a|q,P,A_{\prec a})$. Therefore, the second term in the summation in Eq.\ref{eq: training_objective}, the discrimination objective, optimizes the model to assign a higher faithfulness score to $a$ over $a^{'}$.

Note that the above training objective is applied throughout the whole training, but the training sets $\{(q, A_{\prec a}, a, a^{'})\}$, which are constructed using the latest trained model, vary across different training iterations. \S\ref{sec:data construction} will elaborate the process of data construction.

\subsection{Iterative Data Construction}
\label{sec:data construction}
To enable sentence-level optimization with Eq.\ref{eq: training_objective}, the training instances must include 1) a contrastive sentence pair $(a, a^{'})$ that share a common prefix $A_{\prec a}$; and 2) the relative relation in the faithfulness degree of the pair, i.e., $S_{a} > S_{a^{'}}$. Based on this, we design following data construction methods for pre-stage (first iteration) and self-evolving stages (remaining iterations), respectively.

\subsubsection{Pre-Stage}
\label{sec: per-stage-data-construction}
Initially, the pre-trained model $\theta_{0}$ lacks sufficient self-scoring capability to directly differentiate answer candidates in varying faithfulness
among self-generations. To address this, we construct the pre-stage training dataset $\mathcal{D}_0$, in which sentences from ground-truth answers $a^*$ serve as target instances, and model-generated responses, produced without access to evidence passages (i.e., conditioned only on the question and answer prefix), serve as negative instances as shown in Fig.\ref{fig: overview}(a). 
Specifically, for each target sentence $a$ in a ground-truth answer, we obtain corresponding negative sentence $a^{'}$ as
$a^{'} \sim \mathcal{P}_{\theta_{0}}(* | q, A_{\prec a})$.
Although $a^{'}$ are likely to lack sufficient faithfulness due to the absence of supporting passages, it remains possible for $\theta_{0}$ to answer the question correctly given its extensive pre-training on massive corpora. Consequently, we implement a heuristic negative sample filtering process for quality control with model $\theta_0$ (see details App. \ref{sec:appendix-method}), trying to ensure that the faithfulness score of the positive instance exceeds that of its negative counterpart, i.e., $S_{a} > S_{a^{'}}$.



\subsubsection{Self-evolving Stages}
\label{sec: self-evolving stages}
\noindent\textbf{Tree-structured Sampling} For self-evolving stages, we no longer use the ground-truth answers on the training set. Instead, given the question and passages on the training set, we sample and score answer sentence pairs with the latest model. To efficiently obtain diverse and high-quality sentence pairs, we devise a tree-structured sampling method, the operation process of which can be illustrated with an n-ary tree as shown in Fig.\ref{fig: overview}(b).

\noindent The root node of the tree represents an empty string. Each non-root node indicates a sentence $a$, which is sampled based on its all prefix sentences along the path from the root to its parent node. Therefore, each path from the root to a leaf node constitutes a complete answer and the node in $i$-th layer $a_i$ is thus the $i-$th sentence in the answer. Formally, $a_i \sim \mathcal{P}_{\theta}(* |q, P, a_1, ..., a_{i-1})$. We sample $n$ distinct sentences for each generation, finally leading to an $n$-ary tree (using $n=2$ as an example in Fig.\ref{fig: overview}(b)). During the expansion of the tree, a path will be terminated once $[eos]$ is output. The tree expands layer by layer until it reaches the specified size or all paths are terminated.

\noindent\textbf{Sentence Pairs Selection} We then construct contrastive sentence pairs from the generated tree. Specifically, we evaluate all the terminated paths on the tree, and select the one that achieves the highest accuracy, measured by Exact Match (EM), against the gold answer. Each node $a$ along the path, together with its sibling node $a'$, will compose a contrastive sentence pair $(a, a')$, if $S_{a} > S_{a^{'}}$. Note that the sentences of sibling nodes must have a common prefix.

\subsection{Hierarchical Inference}
\label{sec:inference}

When making inference with the trained model, one can use some standard decoding methods like greedy search. However, this will make the model's scoring ability idle. To make full use the ability of sentence-level scoring, we propose hierarchical inference for answer generation. This method utilizes the model's scoring ability to carry out sentence-level search during generation.

The hierarchical inference is a two-level inference method. The inner one is token-level inference, used to determine the tokens within a sentence. It is similar to the standard decoding methods which decode text token by token, so conventional methods like beam search or top-$k/p$ sampling can be applied to the token-level inference. 


More importantly, we also want to optimize answers in sentence level during inference. We view a complete sentence as a step and generate multiple candidate sentence iteratively. Based on their faithfulness scores, we then utilize beam search algorithm to determine the sentence at each step. In details, a fixed number $N$ of beams are maintained throughout the generation process. When generating the $i$-th sentence, for each beam, based on $q$, $P$ and $A_{<i}$, $M$ (beam width) candidate sentences are sampled through token-level inference. Among the $NM$ continuations, the top $N$ beams with the highest length-normalized faithfulness scores, defined as $\frac{\sum_{j=1}^{i}{S_{a_j}}}{i}$, are selected for the next step of generation. Again, each of these selected beams are expanded by sampling $M$ next sentences. The process will stop once all the selected beams end with $[eos]$ or the maximum search depth is exceeded. Finally, the beam with the highest length-normalized faithfulness score will be returned as the answer.

\section{Experiments}
\subsection{Datasets \& Metrics}
We conduct experiments on two benchmarks: 1) \textbf{ASQA} \citep{stelmakh-etal-2022-asqa}, a long-form factoid question answering dataset derived from AmbigQA \citep{min-etal-2020-ambigqa} , which includes crowd-sourced, paragraph-long answers to ambiguous questions; and 
2) \textbf{ConFiQA} \citep{bi2024contextdpoaligninglanguagemodels}, a dataset designed for retrieval-augmented generation (RAG) scenarios, incorporating knowledge conflicts through counterfactual passages to assess context-faithfulness.
For all reported results, we use the same retrieval results with GTR \citep{ni2021largedualencodersgeneralizable} as dense retriever for ASQA dataset following \citet{gao-etal-2023-enabling, aly-etal-2024-learning}, and the counterfactual contexts provided by \citet{bi2024contextdpoaligninglanguagemodels} for ConFiQA.
We train our model exclusively on ASQA (in-domain) and  evaluate it on ConFiQA (out-of-domain) to assess its generalization and adaptation ability. See data statistics in App.\ref{sec:appendix-data}.

We follow previous literature \citep{aly-etal-2024-learning, gao-etal-2023-enabling} to measure correctness (via EM Recall and hit rate), 
and faithfulness (via an NLI-trained T5-11B model \citep{honovich-etal-2022-true} and AlignScore \citep{zha-etal-2023-alignscore}). In the case of ConFiQA, the ground-truth answer is a single entity name, along with its aliases. Consequently, EM Recall is equivalent to the hit rate. 

\subsection{Implementation Details}
\noindent\textbf{Model Training}: $\lambda$ in Eq.\ref{eq: training_objective} is set to 0.5. We train the model for 3 iterations, and each iteration takes one epoch. \texttt{Llama-3.1-8b} \citep{grattafiori2024llama3herdmodels} is used as the base model. Refer to App.\ref{sec:appendix-training} for more training details.

\noindent\textbf{Data construction}: During tree-structured sampling at self-evolving stages, for each continuation, $n=3$ sentences are sampled with random sampling ($p$=0.9, temperature=1).

\noindent\textbf{Hierarchical Decoding}: For sentence-level inference, both $N$ and $M$ are 3. We thus use beam search with beam width as 3 for token-level inference and produce 3 sentences at each step. The maximum number of steps for sentence-level inference is set to 6.

\subsection{Baselines}
\label{sec: baselines}
To provide a comprehensive evaluation, we compare \text{GenDiE} with two types of baselines.

\noindent\textit{Training-free Approaches} 
1) \textbf{In-context Prompting} generates answers with two demonstration examples by \texttt{gpt-4o} \citep{openai_blog_gpt4o_2024} and \texttt{Llama-3.1-8b-Instruct} \citep{grattafiori2024llama3herdmodels}. The complete prompt is shown in App.\ref{sec:appendix-prompt}.
2) \textbf{CAD} \citep{shi-etal-2024-trusting} uses contrastive decoding to amplify the difference in output probabilities with and without context, reinforcing the model's attention to input context during inference. The backbone model is \texttt{Llama-3.1-8b} with standard fine-tuning described in baseline-4.
3) \textbf{Extractive Sentence Selection} uses embedding models \texttt{stella\_en\_1.5B\_v5}\footnote{\url{https://huggingface.co/NovaSearch/stella_en_1.5B_v5}} and \texttt{instructor-large}\footnote{\url{https://huggingface.co/hkunlp/instructor-large}} to select top-K relevant passage sentences according to the question as final answer, where K depends on \text{GenDiE} for fair comparison.

\noindent\textit{Training-based Approaches} 
4) \textbf{Standard SFT} directly fine-tune LLMs to replicate ground-truth answers.
5) \textbf{\text{GenDiE\textsubscript{answer-level}}} is a variant of our method that using the same training paradigm but optimize in \textit{answer} level instead of \textit{sentence} level. In this setting, complete answers instead of separate sentences, are sampled and scored by the model to construct contrastive answer pairs.
6) \textbf{\text{GenDiE\textsubscript{gold-answer}}} always uses sentences from ground-truth answers as target instead of self-generations with highest self-scored values, which is the setting used in the first iteration of \text{GenDiE}. 

\section{Main Results}
\label{sec: main results}
\begin{table*}[]
\centering
\scalebox{0.67}{
\begin{tabular}{clccccccc}
\toprule
\multicolumn{1}{l}{} &  & \multicolumn{4}{c}{ASQA} & \multicolumn{3}{c}{ConFiQA} \\
\multicolumn{1}{l}{} &  & \multicolumn{2}{c}{Faithfulness} & \multicolumn{2}{c}{Correctness} & \multicolumn{2}{c}{Faithfulness} & Correctness \\ \midrule
Type & Method (Decoding) & AlignScore & T5NLI & EM Rec. & Hit & AlignScore & T5NLI & Hit \\ \midrule \midrule
\multirow{5}{*}{Training-Free} & \multicolumn{1}{l|}{In-context Promping: \texttt{gpt-4o}} & 75.11 & 71.52 & \textbf{47.21} & \multicolumn{1}{c|}{18.57} & 44.56 & 34.11 & 35.96 \\
 & \multicolumn{1}{l|}{In-context Promping: \texttt{llama3.1-8b}} & 76.90 & 70.39 & 40.30 & \multicolumn{1}{c|}{14.35} & 50.67 & 30.78 & 55.14 \\
 & \multicolumn{1}{l|}{CAD: \texttt{llama3.1-8b} Standard SFT, greedy} & 72.07 & 66.18 & 42.72 & \multicolumn{1}{c|}{17.61} & \multicolumn{1}{c}{77.46} & \multicolumn{1}{c}{52.79} & \multicolumn{1}{c}{49.38} \\
 & \multicolumn{1}{l|}{Extractive Sentence Selection: \texttt{stella-1.5b}} & -- & -- & 41.00 & \multicolumn{1}{c|}{17.19} & -- & -- & 42.31 \\
 & \multicolumn{1}{l|}{Extractive Sentence Selection: \texttt{instructor-large}} & -- & -- & 36.58 & \multicolumn{1}{c|}{14.45} & -- & -- & 39.17 \\ \midrule
\multirow{4}{*}{Training-based} & \multicolumn{1}{l|}{Standard SFT: \texttt{llama3.1-8b}, greedy} & 65.27 & 57.54 & 41.93 & \multicolumn{1}{c|}{16.35} & 44.94 & 40.74 & 34.58 \\
 & \multicolumn{1}{l|}{Standard SFT: \texttt{llama3.1-8b}, beam3} & 73.88 & 66.47 & 44.64 & \multicolumn{1}{c|}{19.51} & 66.03 & 62.90 & 59.68 \\
 & \multicolumn{1}{l|}{\text{GenDiE\textsubscript{answer-level}}, greedy} & 64.48 & 56.51 & 43.13 & \multicolumn{1}{c|}{19.20} & \multicolumn{1}{c}{69.81} & 66.17 & 77.91 \\
 & \multicolumn{1}{l|}{\text{GenDiE\textsubscript{gold-answer}}, greedy} & 71.66 & 64.76 & 42.88 & \multicolumn{1}{c|}{17.62} & \multicolumn{1}{c}{64.72} & 59.89 & 65.27 \\ \midrule
\multirow{3}{*}{Ours} & \multicolumn{1}{l|}{Greedy search} & 73.87 & 69.16 & 43.61 & \multicolumn{1}{c|}{18.57} & \multicolumn{1}{c}{72.32} &  \underline{70.17} & 73.72 \\
 & \multicolumn{1}{l|}{Vanilla beam search (beam3)} &  \underline{82.30} &  \underline{79.42} & 43.71 & \multicolumn{1}{c|}{ \underline{18.88}} & \multicolumn{1}{c}{ \underline{78.54}} & 69.33 &  \underline{80.95} \\
 & \multicolumn{1}{l|}{Hierarchical inference (beam3-beam3)} & \textbf{84.90} & \textbf{82.03} &  \underline{45.75} & \multicolumn{1}{c|}{\textbf{21.52}} & \multicolumn{1}{c}{\textbf{80.73}}  & \textbf{80.69} & \textbf{84.63} \\ \bottomrule
\end{tabular}
}

\caption{Performance results of different methods on ASQA (in-domain) and ConFiQA (out-of-domain) benchmarks. \textbf{Bold} and  \underline{underline} numbers denote the best and second-best performance. Note that \texttt{Extractive Sentence Selection} directly uses input passage sentences as answers, making its faithfulness inherently 100\%. Consequently, we denote its faithfulness as ``--'' to indicate that faithfulness evaluation is not applicable when compared to other generative approaches.}
\label{tab: main-table}
\end{table*}

\text{GenDiE} with greedy search outperforms most of training-free and training-based baselines in various metrics. Furthermore, even with the same checkpoint, our approach earns significant boost by using hierarchical inference over greedy search or vanilla beam search, demonstrating the benefit of sentence-level search during inference.

For training-free methods, prompting with \texttt{gpt-4o} performs well on ASQA dataset, achieving the best EM Recall, mainly due to its extensive world knowledge obtained during pretraining. However, \texttt{gpt-4o} faces challenges in knowledge conflict scenarios on ConFiQA dataset, as it heavily relies on its own knowledge rather than the provided passages. This observation aligns with the findings reported by \citet{bi2024contextdpoaligninglanguagemodels}.
A similar pattern is observed in prompting with \texttt{Llama}, implying that simply with prompting-based method, the models often disregard those external knowledge that conflicts with their parametric knowledge. 
Extractive Sentence Selection selects the most relevant sentences as answers (and thereby enjoys high faithfulness), but the concatenation of separated sentences, rather than free-form answer generation, often results in low correctness and readability due to its inflexibility. 

Focusing on training-based methods, even with the same training data, \text{GenDiE\textsubscript{gold-answer}} outperform Standard SFT, especially in terms of faithfulness, demonstrating the efficacy of our multi-task training objective. This suggests that the additional training with discrimination objective can also contribute to faithful generation. Furthermore, \text{GenDiE} with greedy search is superior to \text{GenDiE\textsubscript{gold-answer}} across most metrics, which can be attributed to the iterative update of training data (more discussion in \S\ref{section: Effectiveness of Self-Evolving}). However, \text{GenDiE\textsubscript{answer-level}}, which also underwent data update, shows lower faithfulness than Standard SFT on ASQA. This discrepancy between above two comparisons highlights the necessity of sentence-level operation for the data update, as evaluating faithfulness at sentence level is more feasible than at answer level (\S\ref{section: Effectiveness of Sentence-Level Optimization}).

\section{Analysis}
In this section, we conduct ablation studies on the key components of our approach to evaluate their impact. Unless otherwise specified, vanilla greedy search is employed for decoding in all experiments within this section. Details of some variants in experiments are in \S\ref{sec: baselines}.

\subsection{Effectiveness of Self-Evolving}
\label{section: Effectiveness of Self-Evolving}
\begin{figure}[ht]
\includegraphics[width=1\linewidth]{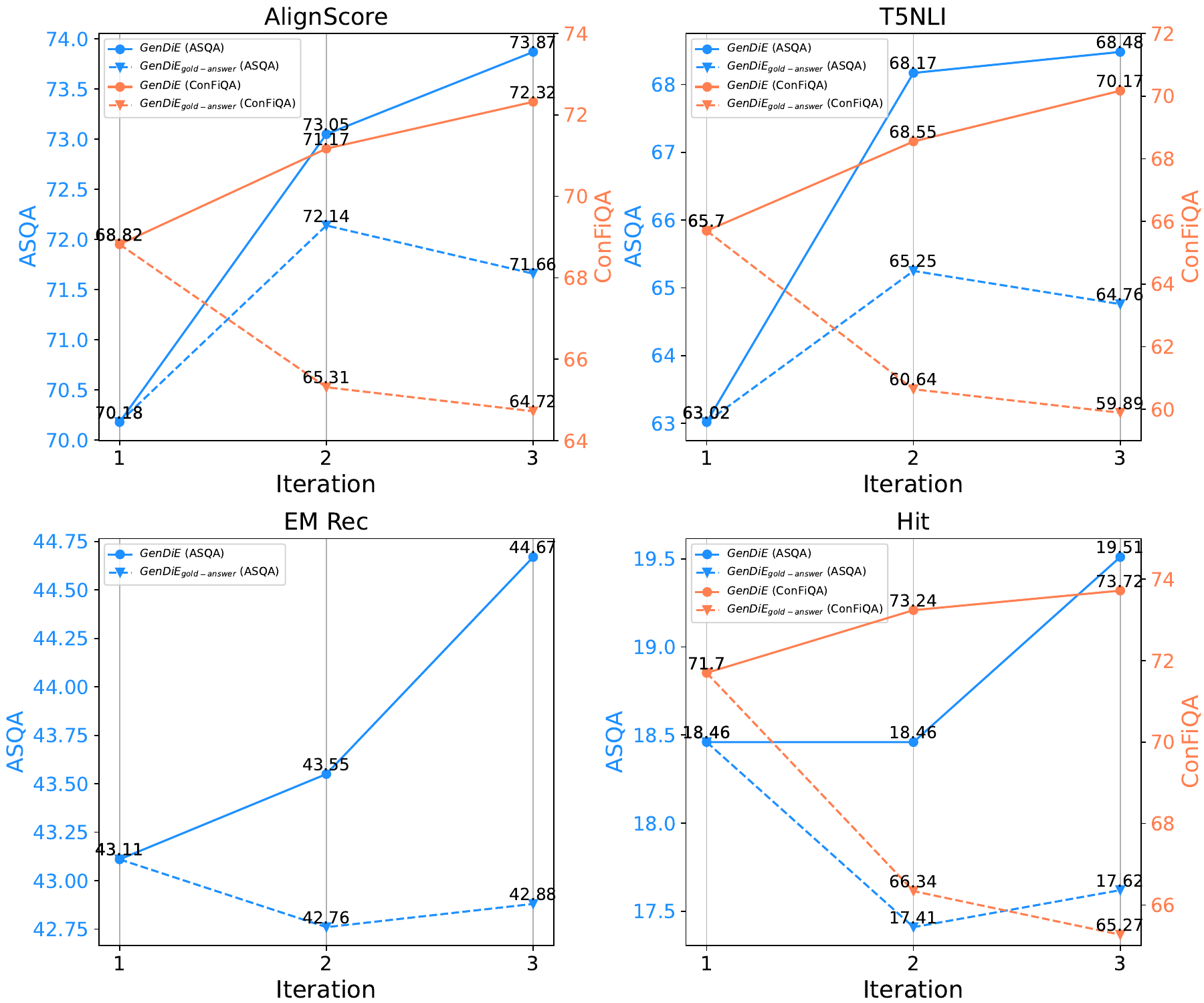}
\caption{Performance comparisons between \text{GenDiE} and \text{GenDiE\textsubscript{gold-answer}} across iterations.}
\label{fig: ours&gold}
\end{figure}
Our approach trains the model with continuously updated data. To investigate the effect of this paradigm, we take a closer look into the performance variation of \text{GenDiE} and \text{GenDiE\textsubscript{gold-answer}} across different iterations.

As shown in Fig.\ref{fig: ours&gold}, with the training progressing, for all metrics, \text{GenDiE} shows continuous improvement on both benchmarks and outperforms \text{GenDiE\textsubscript{gold-answer}} consistently across iterations. Notably, the gap in faithfulness (measured by both AlignScore and T5NLI) between two methods becomes wider from the second to the third iteration. This comparative result underscores the advantage of self-evolving through iterative data update. On ConFiQA, \text{GenDiE\textsubscript{gold-answer}} even experiences declines across various metrics but \text{GenDiE} still gets improved, implying that self-evolving could help alleviate negative effects of out-of-domain settings.

\subsection{Effectiveness of Self-Scoring}
\begin{table}[]
\scalebox{0.7}{
\begin{tabular}{llcccc}
\hline
Dataset & \multicolumn{1}{c}{Checkpoint} & AlignScore & T5NLI & EM Rec. & Hit \\ \midrule \midrule
\multirow{3}{*}{ASQA} & \text{GenDiE\textsubscript{iter1}} & 70.18 & 63.02 & 43.11 & 18.46 \\
 & \text{GenDiE\textsubscript{iter2}} & 73.05 & 68.17 & 43.55 & 18.46 \\
 & \text{GenDiE\textsubscript{T5}} & 73.80 & 70.31 & 42.30 & 17.09 \\ \hline
\multirow{3}{*}{ConFiQA} & \text{GenDiE\textsubscript{iter1}} & 68.82 & 65.70 & - & 71.70 \\
 & \text{GenDiE\textsubscript{iter2}} & 71.17 & 68.55 & - & 73.24 \\
 & \text{GenDiE\textsubscript{T5}} & 71.73 & 69.16 & - & 72.69 \\ \hline
\end{tabular}}
\caption{Performance comparison between \text{GenDiE\textsubscript{iter2}} and \text{GenDiE\textsubscript{T5}}, both of which are trained from \text{GenDiE\textsubscript{iter1}}.}
\label{tab: ours&t5}
\end{table}
For data construction in self-evolving stage (\S \ref{sec: self-evolving stages}), \text{GenDiE} relies the built-in scoring component to assess the faithfulness degrees of sentences, which is necessary for enabling self-evolving. To study the effectiveness of the self-scoring component, we replace it with the NLI-trained T5-11B model for evaluating sentence faithfulness during data construction. The NLI-trained T5 is also used as the evaluation tool to measure faithfulness in previous experiments. With these self-generated but T5-scored data, we train \text{GenDiE\textsubscript{T5}} from \text{GenDiE\textsubscript{iter1}} checkpoint for one iteration, employing the same training objective (Eq.\ref{eq: training_objective}). We then compare it with \text{GenDiE\textsubscript{iter2}}. Note that two methods score and select contrastive pairs from the same collection of sentence pairs sampled from \text{GenDiE\textsubscript{iter1}}. We assess their performances after just one iteration of training, in order to exclude the effect brought by the different self-generated training data at following iterations.

\text{GenDiE\textsubscript{iter2}} exhibits slightly lower T5NLI scores than \text{GenDiE\textsubscript{T5}}. This is expected since \text{GenDiE\textsubscript{T5}} is trained with the direct supervision of T5NLI scores, while \text{GenDiE} relies on the trained self-scoring component. Nevertheless, \text{GenDiE\textsubscript{iter2}} still shows a comparable level of faithfulness on both benchmarks, highlighting the reliable role of the self-scoring component in assessing the faithfulness of sentences when constructing sentence pairs. Also note that \text{GenDiE\textsubscript{T5}} incurs a degradation in correctness, compared with \text{GenDiE\textsubscript{iter1}}.

\subsection{Effectiveness of Sentence-Level Optimization}
\label{section: Effectiveness of Sentence-Level Optimization}
\label{section: Effectiveness of Self-Evolving}
\begin{figure}[ht]
\includegraphics[width=1\linewidth]{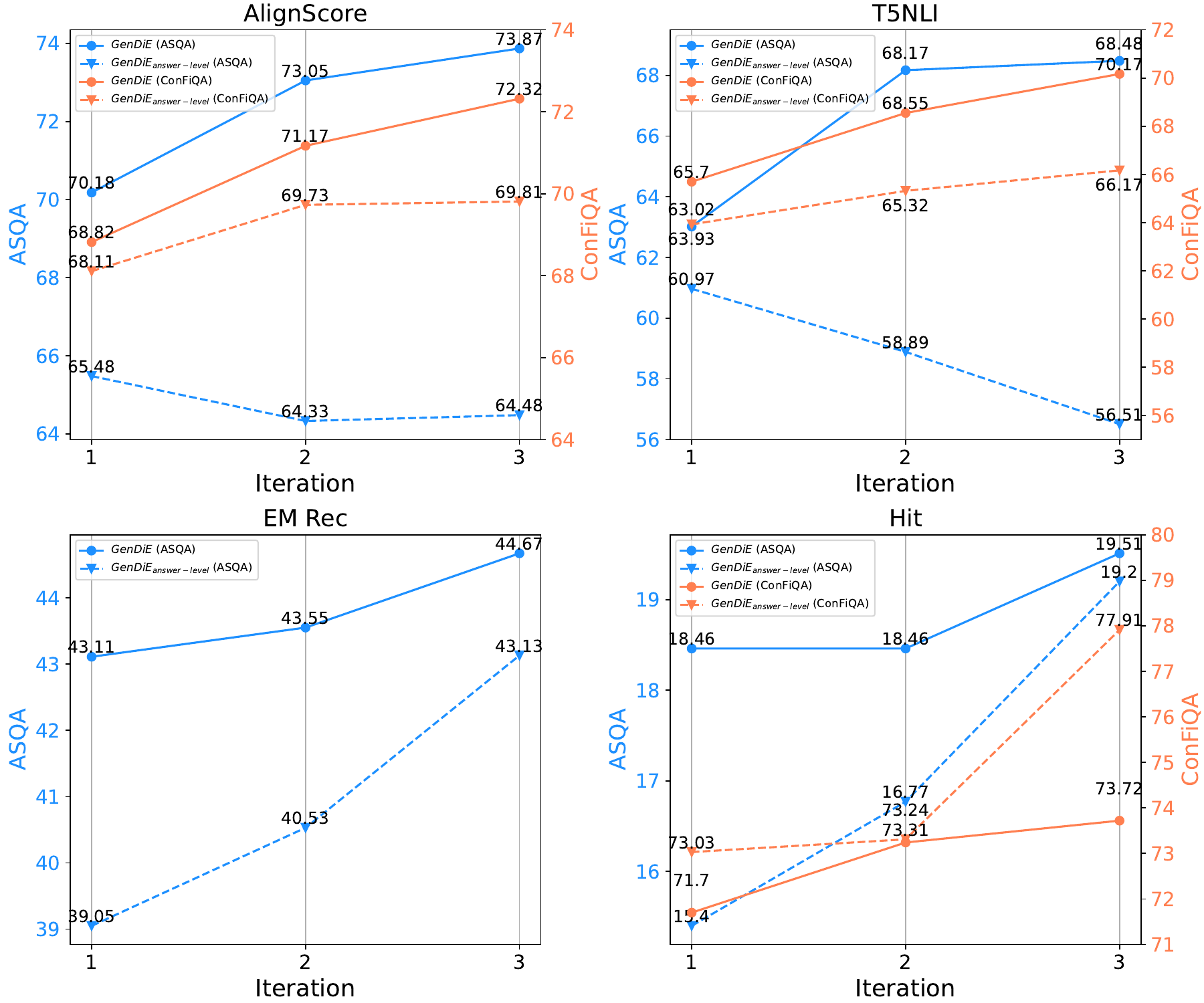}
\caption{The comparisons between \text{GenDiE} and \text{GenDiE\textsubscript{answer-level}} across iterations.}
\label{fig: ours&ans}
\end{figure}
The experiment in \S\ref{sec: main results} demonstrates the significant improvement achieved by hierarchical inference. Besides enabling hierarchical inference, we would like to explore whether sentence-level optimization would contribute to model training. With this aim, we compare the performances between \text{GenDiE} and \text{GenDiE\textsubscript{answer-level}}.

Fig.\ref{fig: ours&ans} plots the changes in the models' performances across different iterations. \text{GenDiE\textsubscript{answer-level}} displays opposite patterns in faithfulness on two benchmarks: it shows consistent improvement on ConFiQA but a decline on ASQA. This is primarily because the questions on ASQA typically demand longer and more detailed answers than on ConFiQA. On ConFiQA, the answer to a question can be an entity, while ASQA often requires paragraph-long answers. \text{GenDiE\textsubscript{answer-level}}, which lacks fine-grained supervision during training, underperforms in yielding faithful long-answers. Moreover, as evaluating faithfulness at answer level is more challenging, it is more difficult to ensure the reliability of self-scored contrastive answer pairs, leading to performance degradation during the self-evolving process.
In comparison, \text{GenDiE} demonstrates increasing faithfulness and consistently outperforms \text{GenDiE\textsubscript{answer-level}}, substantiating the effectiveness of sentence-level optimization. 

Tab. \ref{tab: case-study} shows a comparative case. The generation by \text{GenDiE\textsubscript{answer-level}} is only partially faithful to the passages, with some sentences not being supported by the passages. While \text{GenDiE} provides a fully faithful and coherent answer.

\begin{table}[ht]
\scalebox{0.62}{
\begin{tabular}{p{0.75\textwidth}}
\toprule
\textbf{[Question]}\\ 
When was the first Apple  iPhone made?\\
\textbf{[Passages]}\\ 
The iPhone (colloquially known as the iPhone 2G after 2008) is the first smartphone model designed and marketed by Apple Inc. After years of rumors and speculation, it was officially announced on January 9, 2007, and was later released in the United States on June 29, 2007. ... The original iPhone was introduced by Steve Jobs on January 9, 2007 in a keynote address at the Macworld Conference \& Expo held in Moscone West in San Francisco, California. ...\\ 
\textbf{[\text{GenDiE} generation]}\\ 
The first iPhone was released on June 29, 2007. It was announced by Steve Jobs on January 9, 2007 at the Macworld Conference \& Expo held in Moscone West in San Francisco, California.\\
\textbf{[\text{GenDiE\textsubscript{answer-level}} generation]}\\ 
The first iPhone was released on June 29, 2007. \textcolor{red}{It was a revolutionary device that combined the features of a smartphone, a music player, and a camera into one sleek and stylish device. The iPhone was a game-changer in the mobile phone industry and paved the way for the development of smartphones as we know them today.}\\ 
\bottomrule
\end{tabular}}
\caption{Generation by \text{GenDiE} and \text{GenDiE\textsubscript{answer}} respectively. The questions are from ASQA. The text in \textcolor{red}{red} denotes the content that can not be supported by the provided passages.}
\label{tab: case-study}
\end{table}

\section{Related Work}

\noindent\textbf{Faithfulness} 
Although large language models (LLMs) have showcased impressive capabilities in various tasks \citep{deepseekai2025deepseekr1incentivizingreasoningcapability, zhang-etal-2024-natural, minaee2024largelanguagemodelssurvey}, they are often criticized for generating outputs that deviate from the provided contents, a phenomenon often termed \textit{faithfulness hallucination} \citep{huang-etal-2024-training, zhang2023sirenssongaiocean}.  This issue is particularly pronounced when \textit{knowledge conflict} exists between model's parametric memory and the external evidence, as LLMs may overly rely on their internal priors \citep{jin2024tugofwarknowledgeexploringresolving}. Many approaches have been proposed to improve contextual faithfulness of LLMs. CAD \citep{shi-etal-2024-trusting} leverages contrastive decoding to amplify the difference in output probabilities with and without context, reinforcing the model's attention to input context during inference. 
Self-RAG \citep{asai2024selfrag} trains models to selectively retrieve knowledge and reflect on retrieved information. \citet{luo2023sailsearchaugmentedinstructionlearning} introduces search-augmented instruction learning to ground LLM's generation on search results.
\citet{bi2024contextdpoaligninglanguagemodels} aligns LLMs through DPO \citep{rafailov2024directpreferenceoptimizationlanguage} with constructed faithful and stubborn responses.
While effective, these approaches either intervene LLMs only at inference without addressing inherent limitations \citep{zhou-etal-2023-context}, or fail to equip LLMs with self-judging abilities to explore the potential of self-improvement. Additionally, those answer-level optimization may overlook unfaithful details \citep{lai2024stepdpostepwisepreferenceoptimization}, particularly in long-form generation. In contrast, we propose a novel self-evolving framework with sentence-level optimization, enabling LLMs to enhance context faithfulness through self-generated and self-scored data, fostering iterative self-improvement. Another line of research aims to enhance the accuracy of passage citation in model-generated texts \citep{gao-etal-2023-enabling,huang-etal-2024-training,ye-etal-2024-effective,aly-etal-2024-learning}, thereby increasing the trustworthiness of LLMs. However, these studies primarily focus on citation quality rather than improving the overall answer faithfulness to the input contexts. 


\noindent\textbf{Self-Evolving}
The field of self-evolving mechanisms for large language models (LLMs) is gaining traction as researchers seek to enhance model capabilities beyond current limitations. Self-evolving allows LLMs to autonomously improve and adapt to complex tasks without heavy reliance on human supervision. \citet{huang2022large} illustrates how LLMs can refine reasoning through self-generated rationale-augmented answers, thereby deepening their explanatory capabilities. Self-Align \citep{Sun2023PrincipleDrivenSO} proposes a principle-driven self-alignment model, trained from scratch and requiring little human annotation through self-generated data. Moreover, Self-Rewarding Language Models \citep{yuan2024self} and MathShepherd \citep{wang2024math} present mechanisms where models self-assign high-quality rewards, facilitating their own learning processes. Self-Evolved Reward Learning \citep{Huang2024SelfEvolvedRL} trains the reward model itself using the selected self-labeled  data. Similarly, our approach employs a iterative self-training framework, allowing models to self-generate, self-score, and self-evolve through multiple training iterations. 

\section{Conclusion}
We introduce \text{GenDiE}, a self-evolved approach for enhancing context-faithful generation. 
A distinctive feature of \text{GenDiE} is its capability for generating-then-self-scoring, which facilitates the model's self-evolution via iterative updates to the training data. Additionally, \text{GenDiE} functions at the sentence level, enabling fine-grained control over faithfulness.
Experiments on benchmarks of long-form QA and counterfactual QA show that \text{GenDiE} achieves superior performances in both faithfulness and correctness over various baselines. We also verify the effectiveness of the key designs of \text{GenDiE}. \text{GenDiE} demonstrates promise in building self-evolved and trustworthy RAG system.

\section*{Limitations}
\text{GenDiE} demonstrates notable performances on the benchmarks of long-form QA and counterfactual QA, which are two question-answering tasks with most intensive demand of faithful generation. Nevertheless, we need to show \text{GenDiE} can generalize to other QA tasks that require fast-changing world knowledge. 
Second, although we study the efficacy of multi-task training in \S\ref{sec: main results}, due to computational constraints, we did not conduct investigation of how other training objective for multi-task training, other than the ORPO-based one used in the paper, might affect the performance outcomes. our primary focus lies in achieving self-evolution through models endowed with generative and discriminative capabilities. Delving deeper into the effects of various multi-task training objectives presents a promising opportunity for advancing self-evolution
Lastly, while hierarchical inference significantly enhances performance, it also introduces additional computational overhead compared to standard inference methods, stemming from the multiple generations and explorations required for each sentence. Further efforts to reduce this overhead can be pursued through algorithm optimization. Nevertheless, as observed in other research on LLM reasoning, we should recognize that this test-time scaling yields benefits that substantially outweigh the associated overhead.

\section*{Ethics Statement}
While our framework significantly enhances LLMs' ability to generate contextually faithful responses through self-evolving sentence-level optimization, we emphasize critical ethical considerations. Improved faithfulness scoring does not inherently guarantee factual correctness, as even self-scored "faithful" propositions may inherit biases or contextual omissions from training data. Our work is dedicated to maintaining ethical integrity, ensuring openness in methodology, and advancing the ethical application of AI innovations for societal good.

\bibliography{custom}

\begin{thebibliography}{45}
\providecommand{\natexlab}[1]{#1}

\bibitem[{Aly et~al.(2024)Aly, Tang, Tan, and Karypis}]{aly-etal-2024-learning}
Rami Aly, Zhiqiang Tang, Samson Tan, and George Karypis. 2024.
\newblock \href {https://doi.org/10.18653/v1/2024.acl-long.641} {Learning to generate answers with citations via factual consistency models}.
\newblock In \emph{Proceedings of the 62nd Annual Meeting of the Association for Computational Linguistics (Volume 1: Long Papers)}, pages 11876--11896, Bangkok, Thailand. Association for Computational Linguistics.

\bibitem[{Anthropic(2025)}]{anthropic_blog_clause_3.5_sonnet_2024}
Anthropic. 2025.
\newblock \href {https://www.anthropic.com/news/claude-3-5-sonnet} {Claude 3.5 sonnet}.

\bibitem[{Asai et~al.(2024)Asai, Wu, Wang, Sil, and Hajishirzi}]{asai2024selfrag}
Akari Asai, Zeqiu Wu, Yizhong Wang, Avirup Sil, and Hannaneh Hajishirzi. 2024.
\newblock \href {https://openreview.net/forum?id=hSyW5go0v8} {Self-{RAG}: Learning to retrieve, generate, and critique through self-reflection}.
\newblock In \emph{The Twelfth International Conference on Learning Representations}.

\bibitem[{Bi et~al.(2024)Bi, Huang, Wang, Yang, Zhang, Huang, Mei, Fang, Li, Wei, Deng, Sun, Zhang, and Liu}]{bi2024contextdpoaligninglanguagemodels}
Baolong Bi, Shaohan Huang, Yiwei Wang, Tianchi Yang, Zihan Zhang, Haizhen Huang, Lingrui Mei, Junfeng Fang, Zehao Li, Furu Wei, Weiwei Deng, Feng Sun, Qi~Zhang, and Shenghua Liu. 2024.
\newblock \href {https://arxiv.org/abs/2412.15280} {Context-dpo: Aligning language models for context-faithfulness}.
\newblock \emph{Preprint}, arXiv:2412.15280.

\bibitem[{DeepSeek-AI et~al.(2025)DeepSeek-AI, Guo, Yang, Zhang, Song, Zhang, Xu, Zhu, Ma, Wang, Bi, Zhang, Yu, Wu, Wu, Gou, Shao, Li, Gao, Liu, Xue, Wang, Wu, Feng, Lu, Zhao, Deng, Zhang, Ruan, Dai, Chen, Ji et~al.}]{deepseekai2025deepseekr1incentivizingreasoningcapability}
DeepSeek-AI, Daya Guo, Dejian Yang, Haowei Zhang, Junxiao Song, Ruoyu Zhang, Runxin Xu, Qihao Zhu, Shirong Ma, Peiyi Wang, Xiao Bi, Xiaokang Zhang, Xingkai Yu, Yu~Wu, Z.~F. Wu, Zhibin Gou, Zhihong Shao, Zhuoshu Li, Ziyi Gao, Aixin Liu, Bing Xue, Bingxuan Wang, Bochao Wu, Bei Feng, Chengda Lu, Chenggang Zhao, Chengqi Deng, Chenyu Zhang, Chong Ruan, Damai Dai, Deli Chen, Dongjie Ji, et~al. 2025.
\newblock \href {https://arxiv.org/abs/2501.12948} {Deepseek-r1: Incentivizing reasoning capability in llms via reinforcement learning}.
\newblock \emph{Preprint}, arXiv:2501.12948.

\bibitem[{Dettmers et~al.(2023)Dettmers, Pagnoni, Holtzman, and Zettlemoyer}]{Dettmers2023QLoRAEF}
Tim Dettmers, Artidoro Pagnoni, Ari Holtzman, and Luke Zettlemoyer. 2023.
\newblock \href {https://api.semanticscholar.org/CorpusID:258841328} {Qlora: Efficient finetuning of quantized llms}.
\newblock \emph{ArXiv}, abs/2305.14314.

\bibitem[{Fan et~al.(2019)Fan, Jernite, Perez, Grangier, Weston, and Auli}]{fan-etal-2019-eli5}
Angela Fan, Yacine Jernite, Ethan Perez, David Grangier, Jason Weston, and Michael Auli. 2019.
\newblock \href {https://doi.org/10.18653/v1/P19-1346} {{ELI}5: Long form question answering}.
\newblock In \emph{Proceedings of the 57th Annual Meeting of the Association for Computational Linguistics}, pages 3558--3567, Florence, Italy. Association for Computational Linguistics.

\bibitem[{Fan et~al.(2024)Fan, Ding, Ning, Wang, Li, Yin, Chua, and Li}]{10.1145/3637528.3671470}
Wenqi Fan, Yujuan Ding, Liangbo Ning, Shijie Wang, Hengyun Li, Dawei Yin, Tat-Seng Chua, and Qing Li. 2024.
\newblock \href {https://doi.org/10.1145/3637528.3671470} {A survey on rag meeting llms: Towards retrieval-augmented large language models}.
\newblock In \emph{Proceedings of the 30th ACM SIGKDD Conference on Knowledge Discovery and Data Mining}, KDD '24, page 6491–6501, New York, NY, USA. Association for Computing Machinery.

\bibitem[{Gao et~al.(2023)Gao, Yen, Yu, and Chen}]{gao-etal-2023-enabling}
Tianyu Gao, Howard Yen, Jiatong Yu, and Danqi Chen. 2023.
\newblock \href {https://doi.org/10.18653/v1/2023.emnlp-main.398} {Enabling large language models to generate text with citations}.
\newblock In \emph{Proceedings of the 2023 Conference on Empirical Methods in Natural Language Processing}, pages 6465--6488, Singapore. Association for Computational Linguistics.

\bibitem[{Gao et~al.(2024)Gao, Xiong, Gao, Jia, Pan, Bi, Dai, Sun, Wang, and Wang}]{gao2024retrievalaugmentedgenerationlargelanguage}
Yunfan Gao, Yun Xiong, Xinyu Gao, Kangxiang Jia, Jinliu Pan, Yuxi Bi, Yi~Dai, Jiawei Sun, Meng Wang, and Haofen Wang. 2024.
\newblock \href {https://arxiv.org/abs/2312.10997} {Retrieval-augmented generation for large language models: A survey}.
\newblock \emph{Preprint}, arXiv:2312.10997.

\bibitem[{Grattafiori et~al.(2024)Grattafiori, Dubey, Jauhri, Pandey, Kadian, Al-Dahle, Letman, Mathur, Schelten, Vaughan et~al.}]{grattafiori2024llama3herdmodels}
Aaron Grattafiori, Abhimanyu Dubey, Abhinav Jauhri, Abhinav Pandey, Abhishek Kadian, Ahmad Al-Dahle, Aiesha Letman, Akhil Mathur, Alan Schelten, Alex Vaughan, et~al. 2024.
\newblock \href {https://arxiv.org/abs/2407.21783} {The llama 3 herd of models}.
\newblock \emph{Preprint}, arXiv:2407.21783.

\bibitem[{Hong et~al.(2024)Hong, Lee, and Thorne}]{hong-etal-2024-orpo}
Jiwoo Hong, Noah Lee, and James Thorne. 2024.
\newblock \href {https://doi.org/10.18653/v1/2024.emnlp-main.626} {{ORPO}: Monolithic preference optimization without reference model}.
\newblock In \emph{Proceedings of the 2024 Conference on Empirical Methods in Natural Language Processing}, pages 11170--11189, Miami, Florida, USA. Association for Computational Linguistics.

\bibitem[{Honovich et~al.(2022)Honovich, Aharoni, Herzig, Taitelbaum, Kukliansy, Cohen, Scialom, Szpektor, Hassidim, and Matias}]{honovich-etal-2022-true}
Or~Honovich, Roee Aharoni, Jonathan Herzig, Hagai Taitelbaum, Doron Kukliansy, Vered Cohen, Thomas Scialom, Idan Szpektor, Avinatan Hassidim, and Yossi Matias. 2022.
\newblock \href {https://doi.org/10.18653/v1/2022.dialdoc-1.19} {{TRUE}: Re-evaluating factual consistency evaluation}.
\newblock In \emph{Proceedings of the Second DialDoc Workshop on Document-grounded Dialogue and Conversational Question Answering}, pages 161--175, Dublin, Ireland. Association for Computational Linguistics.

\bibitem[{Hu et~al.(2021)Hu, Shen, Wallis, Allen-Zhu, Li, Wang, and Chen}]{Hu2021LoRALA}
J.~Edward Hu, Yelong Shen, Phillip Wallis, Zeyuan Allen-Zhu, Yuanzhi Li, Shean Wang, and Weizhu Chen. 2021.
\newblock \href {https://api.semanticscholar.org/CorpusID:235458009} {Lora: Low-rank adaptation of large language models}.
\newblock \emph{ArXiv}, abs/2106.09685.

\bibitem[{Huang et~al.(2024{\natexlab{a}})Huang, Fan, Wang, Yang, Zhao, Lin, Lin, Zhang, Rajmohan, and Zhang}]{Huang2024SelfEvolvedRL}
Chenghua Huang, Zhizhen Fan, Lu~Wang, Fangkai Yang, Pu~Zhao, Zeqi Lin, Qingwei Lin, Dongmei Zhang, S.~Rajmohan, and Qi~Zhang. 2024{\natexlab{a}}.
\newblock \href {https://api.semanticscholar.org/CorpusID:273798163} {Self-evolved reward learning for llms}.
\newblock \emph{ArXiv}, abs/2411.00418.

\bibitem[{Huang et~al.(2024{\natexlab{b}})Huang, Wu, Hu, and Wang}]{huang-etal-2024-training}
Chengyu Huang, Zeqiu Wu, Yushi Hu, and Wenya Wang. 2024{\natexlab{b}}.
\newblock \href {https://doi.org/10.18653/v1/2024.acl-long.161} {Training language models to generate text with citations via fine-grained rewards}.
\newblock In \emph{Proceedings of the 62nd Annual Meeting of the Association for Computational Linguistics (Volume 1: Long Papers)}, pages 2926--2949, Bangkok, Thailand. Association for Computational Linguistics.

\bibitem[{Huang et~al.(2022)Huang, Gu, Hou, Wu, Wang, Yu, and Han}]{huang2022large}
Jiaxin Huang, Shixiang~Shane Gu, Le~Hou, Yuexin Wu, Xuezhi Wang, Hongkun Yu, and Jiawei Han. 2022.
\newblock Large language models can self-improve.
\newblock \emph{arXiv preprint arXiv:2210.11610}.

\bibitem[{Jin et~al.(2024)Jin, Cao, Chen, Liu, Jiang, Xu, Li, and Zhao}]{jin2024tugofwarknowledgeexploringresolving}
Zhuoran Jin, Pengfei Cao, Yubo Chen, Kang Liu, Xiaojian Jiang, Jiexin Xu, Qiuxia Li, and Jun Zhao. 2024.
\newblock \href {https://arxiv.org/abs/2402.14409} {Tug-of-war between knowledge: Exploring and resolving knowledge conflicts in retrieval-augmented language models}.
\newblock \emph{Preprint}, arXiv:2402.14409.

\bibitem[{Lai et~al.(2024)Lai, Tian, Chen, Yang, Peng, and Jia}]{lai2024stepdpostepwisepreferenceoptimization}
Xin Lai, Zhuotao Tian, Yukang Chen, Senqiao Yang, Xiangru Peng, and Jiaya Jia. 2024.
\newblock \href {https://arxiv.org/abs/2406.18629} {Step-dpo: Step-wise preference optimization for long-chain reasoning of llms}.
\newblock \emph{Preprint}, arXiv:2406.18629.

\bibitem[{Li et~al.(2024)Li, Zhang, Wu, Luo, Glass, and Meng}]{li2024decodinggraphsfaithfulsound}
Kun Li, Tianhua Zhang, Xixin Wu, Hongyin Luo, James Glass, and Helen Meng. 2024.
\newblock \href {https://arxiv.org/abs/2410.18415} {Decoding on graphs: Faithful and sound reasoning on knowledge graphs through generation of well-formed chains}.
\newblock \emph{Preprint}, arXiv:2410.18415.

\bibitem[{Loshchilov and Hutter(2017)}]{Loshchilov2017DecoupledWD}
Ilya Loshchilov and Frank Hutter. 2017.
\newblock \href {https://api.semanticscholar.org/CorpusID:53592270} {Decoupled weight decay regularization}.
\newblock In \emph{International Conference on Learning Representations}.

\bibitem[{Luo et~al.(2023)Luo, Chuang, Gong, Zhang, Kim, Wu, Fox, Meng, and Glass}]{luo2023sailsearchaugmentedinstructionlearning}
Hongyin Luo, Yung-Sung Chuang, Yuan Gong, Tianhua Zhang, Yoon Kim, Xixin Wu, Danny Fox, Helen Meng, and James Glass. 2023.
\newblock \href {https://arxiv.org/abs/2305.15225} {Sail: Search-augmented instruction learning}.
\newblock \emph{Preprint}, arXiv:2305.15225.

\bibitem[{Min et~al.(2020)Min, Michael, Hajishirzi, and Zettlemoyer}]{min-etal-2020-ambigqa}
Sewon Min, Julian Michael, Hannaneh Hajishirzi, and Luke Zettlemoyer. 2020.
\newblock \href {https://doi.org/10.18653/v1/2020.emnlp-main.466} {{A}mbig{QA}: Answering ambiguous open-domain questions}.
\newblock In \emph{Proceedings of the 2020 Conference on Empirical Methods in Natural Language Processing (EMNLP)}, pages 5783--5797, Online. Association for Computational Linguistics.

\bibitem[{Minaee et~al.(2024)Minaee, Mikolov, Nikzad, Chenaghlu, Socher, Amatriain, and Gao}]{minaee2024largelanguagemodelssurvey}
Shervin Minaee, Tomas Mikolov, Narjes Nikzad, Meysam Chenaghlu, Richard Socher, Xavier Amatriain, and Jianfeng Gao. 2024.
\newblock \href {https://arxiv.org/abs/2402.06196} {Large language models: A survey}.
\newblock \emph{Preprint}, arXiv:2402.06196.

\bibitem[{Nguyen et~al.(2024)Nguyen, Pandit, Purushwalkam, Xu, Chen, Ming, Ke, Savarese, Xong, and Joty}]{nguyen2024sfrragcontextuallyfaithfulllms}
Xuan-Phi Nguyen, Shrey Pandit, Senthil Purushwalkam, Austin Xu, Hailin Chen, Yifei Ming, Zixuan Ke, Silvio Savarese, Caiming Xong, and Shafiq Joty. 2024.
\newblock \href {https://arxiv.org/abs/2409.09916} {Sfr-rag: Towards contextually faithful llms}.
\newblock \emph{Preprint}, arXiv:2409.09916.

\bibitem[{Ni et~al.(2021)Ni, Qu, Lu, Dai, Ábrego, Ma, Zhao, Luan, Hall, Chang, and Yang}]{ni2021largedualencodersgeneralizable}
Jianmo Ni, Chen Qu, Jing Lu, Zhuyun Dai, Gustavo~Hernández Ábrego, Ji~Ma, Vincent~Y. Zhao, Yi~Luan, Keith~B. Hall, Ming-Wei Chang, and Yinfei Yang. 2021.
\newblock \href {https://arxiv.org/abs/2112.07899} {Large dual encoders are generalizable retrievers}.
\newblock \emph{Preprint}, arXiv:2112.07899.

\bibitem[{OpenAI(2024)}]{openai_blog_gpt4o_2024}
OpenAI. 2024.
\newblock \href {https://openai.com/index/hello-gpt-4o/} {Hello gpt-4o}.

\bibitem[{Rafailov et~al.(2024)Rafailov, Sharma, Mitchell, Ermon, Manning, and Finn}]{rafailov2024directpreferenceoptimizationlanguage}
Rafael Rafailov, Archit Sharma, Eric Mitchell, Stefano Ermon, Christopher~D. Manning, and Chelsea Finn. 2024.
\newblock \href {https://arxiv.org/abs/2305.18290} {Direct preference optimization: Your language model is secretly a reward model}.
\newblock \emph{Preprint}, arXiv:2305.18290.

\bibitem[{Shi et~al.(2024)Shi, Han, Lewis, Tsvetkov, Zettlemoyer, and Yih}]{shi-etal-2024-trusting}
Weijia Shi, Xiaochuang Han, Mike Lewis, Yulia Tsvetkov, Luke Zettlemoyer, and Wen-tau Yih. 2024.
\newblock \href {https://doi.org/10.18653/v1/2024.naacl-short.69} {Trusting your evidence: Hallucinate less with context-aware decoding}.
\newblock In \emph{Proceedings of the 2024 Conference of the North American Chapter of the Association for Computational Linguistics: Human Language Technologies (Volume 2: Short Papers)}, pages 783--791, Mexico City, Mexico. Association for Computational Linguistics.

\bibitem[{Stelmakh et~al.(2022)Stelmakh, Luan, Dhingra, and Chang}]{stelmakh-etal-2022-asqa}
Ivan Stelmakh, Yi~Luan, Bhuwan Dhingra, and Ming-Wei Chang. 2022.
\newblock \href {https://doi.org/10.18653/v1/2022.emnlp-main.566} {{ASQA}: Factoid questions meet long-form answers}.
\newblock In \emph{Proceedings of the 2022 Conference on Empirical Methods in Natural Language Processing}, pages 8273--8288, Abu Dhabi, United Arab Emirates. Association for Computational Linguistics.

\bibitem[{Stolfo(2024)}]{stolfo-2024-groundedness}
Alessandro Stolfo. 2024.
\newblock \href {https://doi.org/10.18653/v1/2024.findings-naacl.100} {Groundedness in retrieval-augmented long-form generation: An empirical study}.
\newblock In \emph{Findings of the Association for Computational Linguistics: NAACL 2024}, pages 1537--1552, Mexico City, Mexico. Association for Computational Linguistics.

\bibitem[{Sun et~al.(2023)Sun, Shen, Zhou, Zhang, Chen, Cox, Yang, and Gan}]{Sun2023PrincipleDrivenSO}
Zhiqing Sun, Yikang Shen, Qinhong Zhou, Hongxin Zhang, Zhenfang Chen, David~D. Cox, Yiming Yang, and Chuang Gan. 2023.
\newblock \href {https://api.semanticscholar.org/CorpusID:258479665} {Principle-driven self-alignment of language models from scratch with minimal human supervision}.
\newblock \emph{ArXiv}, abs/2305.03047.

\bibitem[{Wang et~al.(2023)Wang, Liu, Yue, Tang, Zhang, Jiayang, Yao, Gao, Hu, Qi, Wang, Yang, Wang, Xie, Zhang, and Zhang}]{wang2023surveyfactualitylargelanguage}
Cunxiang Wang, Xiaoze Liu, Yuanhao Yue, Xiangru Tang, Tianhang Zhang, Cheng Jiayang, Yunzhi Yao, Wenyang Gao, Xuming Hu, Zehan Qi, Yidong Wang, Linyi Yang, Jindong Wang, Xing Xie, Zheng Zhang, and Yue Zhang. 2023.
\newblock \href {https://arxiv.org/abs/2310.07521} {Survey on factuality in large language models: Knowledge, retrieval and domain-specificity}.
\newblock \emph{Preprint}, arXiv:2310.07521.

\bibitem[{Wang et~al.(2024)Wang, Li, Shao, Xu, Dai, Li, Chen, Wu, and Sui}]{wang2024math}
Peiyi Wang, Lei Li, Zhihong Shao, Runxin Xu, Damai Dai, Yifei Li, Deli Chen, Yu~Wu, and Zhifang Sui. 2024.
\newblock Math-shepherd: Verify and reinforce llms step-by-step without human annotations.
\newblock In \emph{Proceedings of the 62nd Annual Meeting of the Association for Computational Linguistics (Volume 1: Long Papers)}, pages 9426--9439.

\bibitem[{Xie et~al.(2024)Xie, Zhang, Chen, Lou, and Su}]{xie2024adaptive}
Jian Xie, Kai Zhang, Jiangjie Chen, Renze Lou, and Yu~Su. 2024.
\newblock \href {https://openreview.net/forum?id=auKAUJZMO6} {Adaptive chameleon or stubborn sloth: Revealing the behavior of large language models in knowledge conflicts}.
\newblock In \emph{The Twelfth International Conference on Learning Representations}.

\bibitem[{Xu et~al.(2023)Xu, Song, Iyyer, and Choi}]{xu-etal-2023-critical}
Fangyuan Xu, Yixiao Song, Mohit Iyyer, and Eunsol Choi. 2023.
\newblock \href {https://doi.org/10.18653/v1/2023.acl-long.181} {A critical evaluation of evaluations for long-form question answering}.
\newblock In \emph{Proceedings of the 61st Annual Meeting of the Association for Computational Linguistics (Volume 1: Long Papers)}, pages 3225--3245, Toronto, Canada. Association for Computational Linguistics.

\bibitem[{Xu et~al.(2024)Xu, Qi, Guo, Wang, Wang, Zhang, and Xu}]{xu2024knowledgeconflictsllmssurvey}
Rongwu Xu, Zehan Qi, Zhijiang Guo, Cunxiang Wang, Hongru Wang, Yue Zhang, and Wei Xu. 2024.
\newblock \href {https://arxiv.org/abs/2403.08319} {Knowledge conflicts for llms: A survey}.
\newblock \emph{Preprint}, arXiv:2403.08319.

\bibitem[{Ye et~al.(2024)Ye, Sun, Arik, and Pfister}]{ye-etal-2024-effective}
Xi~Ye, Ruoxi Sun, Sercan Arik, and Tomas Pfister. 2024.
\newblock \href {https://doi.org/10.18653/v1/2024.naacl-long.346} {Effective large language model adaptation for improved grounding and citation generation}.
\newblock In \emph{Proceedings of the 2024 Conference of the North American Chapter of the Association for Computational Linguistics: Human Language Technologies (Volume 1: Long Papers)}, pages 6237--6251, Mexico City, Mexico. Association for Computational Linguistics.

\bibitem[{Yu et~al.(2023)Yu, Iter, Wang, Xu, Ju, Sanyal, Zhu, Zeng, and Jiang}]{yu2023generateretrievelargelanguage}
Wenhao Yu, Dan Iter, Shuohang Wang, Yichong Xu, Mingxuan Ju, Soumya Sanyal, Chenguang Zhu, Michael Zeng, and Meng Jiang. 2023.
\newblock \href {https://arxiv.org/abs/2209.10063} {Generate rather than retrieve: Large language models are strong context generators}.
\newblock \emph{Preprint}, arXiv:2209.10063.

\bibitem[{Yuan et~al.(2024)Yuan, Pang, Cho, Sukhbaatar, Xu, and Weston}]{yuan2024self}
Weizhe Yuan, Richard~Yuanzhe Pang, Kyunghyun Cho, Sainbayar Sukhbaatar, Jing Xu, and Jason Weston. 2024.
\newblock Self-rewarding language models.
\newblock \emph{arXiv preprint arXiv:2401.10020}.

\bibitem[{Zha et~al.(2023)Zha, Yang, Li, and Hu}]{zha-etal-2023-alignscore}
Yuheng Zha, Yichi Yang, Ruichen Li, and Zhiting Hu. 2023.
\newblock \href {https://doi.org/10.18653/v1/2023.acl-long.634} {{A}lign{S}core: Evaluating factual consistency with a unified alignment function}.
\newblock In \emph{Proceedings of the 61st Annual Meeting of the Association for Computational Linguistics (Volume 1: Long Papers)}, pages 11328--11348, Toronto, Canada. Association for Computational Linguistics.

\bibitem[{Zhang et~al.(2024{\natexlab{a}})Zhang, Zhang, Long, Xie, Zhang, and Zhang}]{zhang-etal-2024-two}
Longhui Zhang, Yanzhao Zhang, Dingkun Long, Pengjun Xie, Meishan Zhang, and Min Zhang. 2024{\natexlab{a}}.
\newblock \href {https://doi.org/10.18653/v1/2024.findings-acl.706} {A two-stage adaptation of large language models for text ranking}.
\newblock In \emph{Findings of the Association for Computational Linguistics: ACL 2024}, pages 11880--11891, Bangkok, Thailand. Association for Computational Linguistics.

\bibitem[{Zhang et~al.(2024{\natexlab{b}})Zhang, Ge, Luo, Chuang, Gao, Gong, Kim, Wu, Meng, and Glass}]{zhang-etal-2024-natural}
Tianhua Zhang, Jiaxin Ge, Hongyin Luo, Yung-Sung Chuang, Mingye Gao, Yuan Gong, Yoon Kim, Xixin Wu, Helen Meng, and James Glass. 2024{\natexlab{b}}.
\newblock \href {https://doi.org/10.18653/v1/2024.findings-naacl.259} {Natural language embedded programs for hybrid language symbolic reasoning}.
\newblock In \emph{Findings of the Association for Computational Linguistics: NAACL 2024}, pages 4131--4155, Mexico City, Mexico. Association for Computational Linguistics.

\bibitem[{Zhang et~al.(2023)Zhang, Li, Cui, Cai, Liu, Fu, Huang, Zhao, Zhang, Chen, Wang, Luu, Bi, Shi, and Shi}]{zhang2023sirenssongaiocean}
Yue Zhang, Yafu Li, Leyang Cui, Deng Cai, Lemao Liu, Tingchen Fu, Xinting Huang, Enbo Zhao, Yu~Zhang, Yulong Chen, Longyue Wang, Anh~Tuan Luu, Wei Bi, Freda Shi, and Shuming Shi. 2023.
\newblock \href {https://arxiv.org/abs/2309.01219} {Siren's song in the ai ocean: A survey on hallucination in large language models}.
\newblock \emph{Preprint}, arXiv:2309.01219.

\bibitem[{Zhou et~al.(2023)Zhou, Zhang, Poon, and Chen}]{zhou-etal-2023-context}
Wenxuan Zhou, Sheng Zhang, Hoifung Poon, and Muhao Chen. 2023.
\newblock \href {https://doi.org/10.18653/v1/2023.findings-emnlp.968} {Context-faithful prompting for large language models}.
\newblock In \emph{Findings of the Association for Computational Linguistics: EMNLP 2023}, pages 14544--14556, Singapore. Association for Computational Linguistics.

\end{thebibliography}

\appendix
\section{Method Details}
\label{sec:appendix-method}
\subsection{Quality Control} Due to extensive pre-training on massive corpora, LLMs encapsulate substantial world knowledge within their parameters \citep{yu2023generateretrievelargelanguage, wang2023surveyfactualitylargelanguage}. As a result, it is possible that $\theta_{0}$ can answer the question correctly and faithfully even without access to the evidence passages, provided it has encountered similar information during pre-training. To mitigate this, we involve a heuristic negative sample filtering process for quality control. 
Given an input sequence $x$, we denote the (length-normalized) negative log-likelihood (NLL) loss in generating the output sequence $y$ with $m$ tokens as follows:
\begin{equation}
\label{eq: loss}
\begin{split}
\mathcal{L}(y|x)=-\frac{1}{m}\sum_{t=1}^{m}\log P(y_t|x,y_{<t}) \\
\end{split}
\end{equation}
For each candidate training instance $\{(q, A_{\prec a}, a, a^{'})\}$, we compare the NLL loss reduction ratio in generating a candidate with and without input passages to select valid negative sentences:
\begin{equation}
\label{eq: filtering}
\begin{split}
 \frac{\mathcal{L}([A_{\prec a},a]|q,P)-\mathcal{L}([A_{\prec a},a]|q)}{\mathcal{L}([A_{\prec a},a]|q)} < \\
\frac{\mathcal{L}([A_{\prec a}, a^{'}]|q,P)-\mathcal{L}([A_{\prec a}),a^{'}]|q)}{\mathcal{L}([A_{\prec a},a^{'}]|q)}   
\end{split}
\end{equation}
where $[*]$ denotes the concatenation.
The motivation is to ensure that the selected negative samples $a'$ remain less faithful to the input question than the positive answer $a$, even when evidence passages are provided. This filtering process tries to preserve the integrity of our faithfulness relation between positive and negative sentences.

\begin{table}[h]
\centering
\scalebox{0.78}{
\begin{tabular}{lcccc}
\toprule
\multicolumn{1}{l}{} & AlignScore & T5NLI & EM Rec. & Hit   \\ \midrule
non-filtered           & 68.55      & 62.02 & 42.81     & 17.93 \\
filtered             & \textbf{70.46}      & \textbf{63.02} & \textbf{43.11}     & \textbf{18.46} \\ \bottomrule
\end{tabular}}
\caption{Performance comparison for pre-stage with and without filtering on ASQA dataset.}
\label{tab: filter-result}
\end{table}
The effectiveness of pre-stage filtering is reported in Tab. \ref{tab: filter-result}. For each training item, we use model $\theta_0$ to sample six negative sentence candidates, as described in \S\ref{sec: per-stage-data-construction}. After filtering via Eq. \ref{eq: filtering}, we retain at most two of the most negative sentences, ranked by the NLL loss reduction ratio, for \texttt{filtered} setting (some items may contain only one valid negative sentence after filtering). In contrast, the \texttt{non-filtered} setting uses two randomly selected sentences from the six candidates. As demonstrated in the Tab. \ref{tab: filter-result}, heuristic negative sample filtering improves both faithfulness and correctness, with \texttt{filtered} outperforming \texttt{non-filterd} across all metrics.  


\section{Implementation Details}
\subsection{Data Statistics}
\label{sec:appendix-data}
Tab. \ref{tab:data-statistics} shows the statistics of the two benchmarks. For ASQA, we refine the original $4353$ instances by retaining only $3414$ where the top-5 passages contain at least one candidate answer from the given reference list. This ensures that the question is at least partially answerable by the input passages. Truncating the answer sentences yields $14841$ items, from which we retain $11531$ valid items after filtering for model training.
\begin{table}[h]
\centering
\scalebox{0.67}{
\begin{tabular}{ccc}
\toprule
Dataset & Train (orig. ans./kept ans./trunc. sent./final sent. ) & Test \\ \midrule
ASQA    & 4353/3414/14841/11531      & 948            \\
ConFiQA & -         & 18000        
\\ \bottomrule
\end{tabular}}
\caption{Data statistics for two benchmarks.}
\label{tab:data-statistics}
\end{table}

\subsection{Training Details}
\label{sec:appendix-training}
We use AdamW \cite{Loshchilov2017DecoupledWD} optimizer with learning rates of 1e-5 . The learning rates undergo a warmup of $10\%$ of overall training steps, followed by a linear decrease until 0. We utilize quantized LoRA (QLoRA) \cite{Hu2021LoRALA, Dettmers2023QLoRAEF} as the parameter-efficient fine-tuning technique to train the models with an NVIDIA A6000 GPU. Specifically, QLoRA is implemented on the query and value attention matrices within each decoder block, using a fixed rank of 8, a scaling factor of 16, and a dropout rate of 0.05. The model weights are quantized and loaded in 4-bit NormalFloat format.

\section{Prompts}
\label{sec:appendix-prompt}
\begin{table*}
\small
    \centering
    \scalebox{0.85}{
    \colorbox{purple!8}{
    \begin{tabular}{@{}p{17.2cm}}
Instruction: Write an accurate, engaging, fluent and detailed answer for the given question using only the provided search results.
\\ \\
Question: Which is the most rainy place on earth?
\\ \\
Document [1](Title: Cherrapunji): Cherrapunji Cherrapunji (; with the native name Sohra being more commonly used, and can also be spelled Cherrapunjee or Cherrapunji) is a subdivisional town in the East Khasi Hills district in the Indian state of Meghalaya. It is the traditional capital of aNongkhlaw "hima" (Khasi tribal chieftainship constituting a petty state), both known as Sohra or Churra. Cherrapunji has often been credited as being the wettest place on Earth, but for now nearby Mawsynram currently holds that distinction. Cherrapunji still holds the all-time record for the most rainfall in a calendar month for July 1861 and most rain in a year from August 1860 to July 1861, however: it received in \\
Document [2](Title: Cherrapunji): Radio relay station known as Akashvani Cherrapunji. It broadcasts on FM frequencies. Cherrapunji Cherrapunji (; with the native name Sohra being more commonly used, and can also be spelled Cherrapunjee or Cherrapunji) is a subdivisional town in the East Khasi Hills district in the Indian state of Meghalaya. It is the traditional capital of aNongkhlaw "hima" (Khasi tribal chieftainship constituting a petty state), both known as Sohra or Churra. Cherrapunji has often been credited as being the wettest place on Earth, but for now nearby Mawsynram currently holds that distinction. Cherrapunji still holds the all-time record for the most rainfall \\
Document [3](Title: Mawsynram): Mawsynram Mawsynram () is a village in the East Khasi Hills district of Meghalaya state in north-eastern India, 65 kilometres from Shillong. Mawsynram receives one of the highest rainfalls in India. It is reportedly the wettest place on Earth, with an average annual rainfall of 11,872 mm, but that claim is disputed by Lloró, Colombia, which reported an average yearly rainfall of 12,717 mm between 1952 and 1989 and López de Micay, also in Colombia, which reported an annual 12,892 mm per year between 1960 and 2012. According to the "Guinness Book of World Records", Mawsynram received of rainfall in 1985. Mawsynram is located at 25° 18' \\
Document [4](Title: Earth rainfall climatology): Pacific Northwest, and the Sierra Nevada range are the wetter portions of the nation, with average rainfall exceeding per year. The drier areas are the Desert Southwest, Great Basin, valleys of northeast Arizona, eastern Utah, central Wyoming, eastern Oregon and Washington and the northeast of the Olympic Peninsula. The Big Bog on the island of Maui receives, on average, every year, making it the wettest location in the US, and all of Oceania. The annual average rainfall maxima across the continent lie across the northwest from northwest Brazil into northern Peru, Colombia, and Ecuador, then along the Atlantic coast of \\
Document [5](Title: Going to Extremes): in the world. Oymyakon in Siberia, where the average winter temperature is -47 °F (-44 °C). Arica in Chile, where there had been fourteen consecutive years without rain. Fog is the only local source of water. Mawsynram in India, where average annual rainfall is 14 meters, falling within a four-month period in the monsoon season. The rainfall is approximately equal to that of its neighbor Cherrapunji. Dallol in Ethiopia, known as the 'Hell-hole of creation' where the temperature averages 94 °F (34 °C) over the year. In his second series, Middleton visited places without permanent towns, locations where "survival"
\\ \\
Answer: Several places on Earth claim to be the most rainy, such as Lloró, Colombia, which reported an average annual rainfall of 12,717 mm between 1952 and 1989, and López de Micay, Colombia, which reported an annual 12,892 mm between 1960 and 2012. However, the official record is held by Mawsynram, India with an average annual rainfall of 11,872 mm, although nearby town Sohra, India, also known as Cherrapunji, holds the record for most rain in a calendar month for July 1861 and most rain in a year from August 1860 to July 1861.
\\ \\
\\
Question: When did the us break away from england?
\\ \\
Document [1](Title: United States withdrawal from Saudi Arabia): United States withdrawal from Saudi Arabia Beginning during Operation Desert Shield in August 1990, while preparing for the Gulf War, the United States sent a large troop contingent to Saudi Arabia. After the war, remnant troops, primarily U.S. Air Force personnel, augmented by a smaller number of coordinating and training personnel from the U.S. Navy, U.S. Army and U.S. Marine Corps remained in Saudi Arabia under the aegis of Joint Task Force Southwest Asia (JTF-SWA), as part of Operation Southern Watch (OSW). The United Kingdom and France also maintained a small contingent of Royal Air Force and French Air Force \\
Document [2](Title: Decolonization of the Americas): and France has fully "integrated" most of its former colonies as fully constituent "departments" of France. The United States of America declared independence from Great Britain on July 2, 1776 (although the event is now commemorated on July 4, the date when the Declaration of Independence was officially adopted by Congress), in so doing becoming the first independent, foreign-recognized nation in the Americas and the first European colonial entity to break from its mother country. Britain formally acknowledged American independence in 1783 after its defeat in the American Revolutionary War. Although initially occupying only the land east of the Mississippi \\
Document [3](Title: American Revolution): second British army at Yorktown in the fall of 1781, effectively ending the war. The Treaty of Paris was signed September 3, 1783, formally ending the conflict and confirming the new nation's complete separation from the British Empire. The United States took possession of nearly all the territory east of the Mississippi River and south of the Great Lakes, with the British retaining control of Canada and Spain taking Florida. Among the significant results of the revolution was the creation of the United States Constitution, establishing a relatively strong federal national government that included an executive, a national judiciary, and \\
Document [4](Title: Decolonization): accelerate decolonialization and bring an end to the colonial empires of its Western allies, most importantly during the 1956 Suez Crisis, but American military bases were established around the world and direct and indirect interventions continued in Korea, Indochina, Latin America ("inter alia", the 1965 occupation of the Dominican Republic), Africa, and the Middle East to oppose Communist invasions and insurgencies. Since the dissolution of the Soviet Union, the United States has been far less active in the Americas, but invaded Afghanistan and Iraq following the September 11 attacks in 2001, establishing army and air bases in Central Asia. Before \\
Document [5](Title: Decolonization): the responsibility of the United Kingdom (with a copy of the new constitution annexed), and finally, if approved, issuance of an Order of Council fixing the exact date of independence. After World War I, several former German and Ottoman territories in the Middle East, Africa, and the Pacific were governed by the UK as League of Nations mandates. Some were administered directly by the UK, and others by British dominions – Nauru and the Territory of New Guinea by Australia, South West Africa by the Union of South Africa, and Western Samoa by New Zealand. Egypt became independent in 1922,
\\ \\
Answer: The United States took the first step towards gaining independence from Great Britain when it declared independence from Great Britain on July 2, 1776 (although the event is now commemorated on July 4, 1776, the date when the Declaration of Independence was officially adopted by Congress). The Treaty of Paris was later signed on September 3, 1783, formally separating the United States from the British Empire.
\\ \\ \\
Question: \{\texttt{test\_question}\}
\\ \\
Documents: \{\texttt{test\_documents}\}
\\ \\
Answer:
\end{tabular}

}}
    \caption{Prompt used for \texttt{In-context Prompting} approach. 
}
    \label{tab:appendix-prompt}
\end{table*}
Tab. \ref{tab:appendix-prompt} presents the prompt used for \texttt{In-context Prompting} approach over the two benchmarks. Two in-context examples are listed, each consisting of five input passages, which matches the setting of the test questions. The in-context learning examples are sourced from the ASQA dataset, following previous approaches \citep{aly-etal-2024-learning}. We do not use in-domain demonstrations for ConFiQA to simulate the out-of-domain evaluation setting.

\end{document}